\documentclass[sigconf]{acmart-arxiv}
\usepackage{tikz}
\usepackage{pgfplots}
\pgfplotsset{compat=1.18}
\usepackage{array}
\usepackage[table]{xcolor}
\usepackage{subcaption}
\usepackage{url}
\microtypesetup{expansion=false}

\AtBeginDocument{%
  }

\setcopyright{none}
\renewcommand\footnotetextcopyrightpermission[1]{}

\acmConference[CIKM '26]{Proceedings of the 35th ACM International Conference on Information and Knowledge Management}{November 7--11, 2026}{Rome, Italy}
\acmYear{2026}
\copyrightyear{2026}
\acmDOI{}
\acmISBN{}

\begin{document}

\title{GraphLoom: Reliability-Calibrated Graph Evidence Routing for Multimodal KG-RAG}

\author{Zafar Ali}
\affiliation{%
  \institution{School of Computer Science and Engineering, Southeast University}
  \city{Nanjing}
  \country{China}
}
\email{zafar_ali@seu.edu.cn}

\author{Asad Khan}
\affiliation{%
  \institution{School of Computer Science and Engineering, Southeast University}
  \city{Nanjing}
  \country{China}
}
\email{asadkhan@seu.edu.cn}

\author{Aalia Malik}
\affiliation{%
  \institution{School of Computer Science and Engineering, Southeast University}
  \city{Nanjing}
  \country{China}
}
\email{aaliamalik@seu.edu.cn}

\author{Pavlos Kefalas}
\affiliation{%
  \institution{Dashub \& Department of Informatics, Aristotle University}
  \city{Thessaloniki}
  \country{Greece}
}
\email{pavloskefalas@gmail.com}

\renewcommand{\shortauthors}{Ali et al.}

\begin{abstract}
Multimodal retrieval-augmented generation (RAG) systems often rely on long unstructured contexts or aggressively expanded evidence graphs, which can introduce noisy evidence, weaken multi-hop reasoning, and increase unsupported generation. We present \textbf{GraphLoom}, a reliability-calibrated multimodal knowledge-graph RAG framework for compact and faithful evidence routing. Given a question and its associated multimodal input, GraphLoom constructs an instance-level multimodal knowledge graph from grounded scene descriptions, extracted relational triples, and external commonsense knowledge. Instead of injecting all retrieved evidence into the generator, GraphLoom performs reliability-aware subgraph retrieval with bounded expansion and selectively routes high-utility evidence through hierarchical graph memory slots and joint graph--sequence attention in a frozen language model. To improve robustness in complex reasoning settings, GraphLoom further combines interleaved retrieval with budgeted corrective retrieval, enabling adaptive multi-hop evidence refinement under noisy retrieval conditions. We evaluate GraphLoom on \textit{ScienceQA}, \textit{MultiModalQA}, and \textit{OK-VQA}, including large distractor evidence pools that approximate noisy external knowledge retrieval. Experimental results show consistent gains in answer quality and evidence faithfulness over strong multimodal RAG, graph-retrieval, and open-source vision--language baselines, with improved retrieval quality on MultiModalQA and stable performance under noisy evidence pools. Additional analyses using MiniCheck-based verification, human evaluation, and latency profiling show that reliability-calibrated graph evidence routing provides an effective alternative to long-context multimodal evidence injection.
\end{abstract}




\maketitle
\sloppy 
\section{Introduction}

Multimodal question answering requires \emph{grounded} reasoning over heterogeneous evidence, including textual descriptions, tables, images or diagrams, and structured relational knowledge. Knowledge-graph question answering (KG-QA)~\cite{Lan2022ComplexKBQA,jiangunikgqa} and retrieval-augmented generation (RAG)~\cite{chen2022murag,G-retriever} offer complementary strengths: graphs expose explicit relational structure for multi-hop reasoning, while neural retrieval supports evidence access over large and noisy information spaces. However, reliable multimodal KG-assisted QA remains challenging because relevant evidence is often distributed across visual content, textual or tabular context, external knowledge, and inferred relations. This makes the problem not only a reasoning task, but also an evidence retrieval and knowledge-management problem.

Despite rapid progress in vision--language models~\cite{liu2023llava,li2023blip}, current multimodal QA pipelines still face several limitations.
\emph{First}, multimodal knowledge-graph reasoning remains sensitive to incomplete knowledge and noisy or imperfectly fused multimodal evidence, which can impair multi-hop inference~\cite{zheng2023mmkgr}.
\emph{Second}, large or broadly expanded graph contexts can exceed practical context budgets and make evidence selection difficult~\cite{G-retriever}, while multimodal generators remain susceptible to visually ungrounded hallucinations~\cite{Leng_2024_CVPR}.
\emph{Third}, scalable multi-hop retrieval and long-context decoding remain expensive. Large KG search spaces motivate retrieval of compact supporting subgraphs~\cite{jiangunikgqa,G-retriever}, while retaining historical key--value states during decoding incurs substantial memory cost~\cite{xiao2023streamingllm}. Sparse-attention mechanisms can reduce this inference burden~\cite{xiao-etal-2025-efficient}.
\emph{Fourth}, many retrieval and correction pipelines depend on multiple thresholds and confidence scores, yet their robustness under parameter variation is often unclear.
\emph{Finally}, using the same verifier for both answer revision and final faithfulness assessment can risk circular evaluation. These issues are especially important for information and knowledge management settings, where systems must retrieve from noisy evidence pools, remain efficient, and provide credible grounding diagnostics.

To address these limitations, we propose \textbf{GraphLoom}, a reliability-calibrated multimodal KG-RAG framework for controlled evidence retrieval, organization, and injection. Given a question and its associated multimodal input, GraphLoom constructs an instance-level MMKG from available evidence, including grounded scene descriptions generated by \textbf{Qwen3-VL-Instruct}~\cite{bai2025qwenvl3} and relational triples extracted by \textbf{REBEL}~\cite{rebel}. The graph is enriched with visual relation priors derived from \textit{Visual Genome} (VG150)~\cite{krishna2017visualgenome} and commonsense knowledge from \textit{ConceptNet}~\cite{speer2017conceptnet}. A unified multimodal embedder, \textbf{Qwen3-VL-Embedding}~\cite{li2026qwen3vlembedding}, performs task-conditioned subgraph retrieval with bounded expansion, producing compact supporting evidence without unbounded graph traversal. To evaluate scalability beyond a small local graph, we further test retrieval under large distractor evidence pools that simulate noisy external knowledge sources.

For multi-hop questions, GraphLoom performs interleaved retrieval, where intermediate hypotheses trigger an additional bounded retrieval round before decoding~\cite{trivedi2023ircot}. When the retrieved evidence appears insufficient, a lightweight adequacy evaluator triggers at most one budgeted corrective action, following the corrective-retrieval principle of CRAG~\cite{yan2024crag}. GraphLoom considers targeted expansion, query reformulation, or evidence re-scoring as bounded corrective actions. After generation, a verification--revision stage edits unsupported spans using retrieved evidence~\cite{gao2023rarr}. To avoid circular faithfulness claims, we evaluate final grounding with an independent verifier and human validation rather than relying only on the verifier used during revision.

A key design goal of GraphLoom is decoder-time evidence control under transparent computation budgets. Retrieved triples are converted into compact key--value slot memories, called \textsc{HieraSlot}. A reliability-calibrated router estimates the utility and reliability of candidate slots and activates only a small subset at each decoding step. These selected slots, together with multimodal prefix memories, are injected into a frozen \textbf{Llama-3.1-8B-Instruct} decoder through \textsc{KG-JSA++} joint graph--sequence attention. This design supports stepwise grounding without relying on long unstructured prompts or broad evidence expansion. We also distinguish full-pipeline latency from online latency under cached graph construction, making the efficiency profile explicit.

We evaluate GraphLoom on \textsc{ScienceQA}, \textsc{OK-VQA}, and \textsc{MultiModalQA} across multiple-choice and open-ended QA settings. Beyond answer correctness, we analyze retrieval quality, evidence faithfulness, scalability under distractor evidence, threshold robustness, and latency. We compare GraphLoom with multimodal RAG, graph-based retrieval, interleaved/corrective retrieval, and recent vision--language baselines. Our main contributions are:
\begin{itemize}

\item We introduce \textbf{GraphLoom}, a reliability-calibrated multimodal KG-RAG framework that treats evidence retrieval, graph organization, and decoder-time evidence use as a controlled knowledge-management problem.

\item We propose a \textbf{controlled evidence injection interface} that converts retrieved graph evidence into multimodal prefix memories and hierarchical key--value graph slots, enabling compact grounding in a frozen language model.

\item We develop a \textbf{reliability-calibrated slot routing mechanism} with \textsc{KG-JSA++} joint graph--sequence attention, allowing useful evidence slots to compete with sparse self-context during decoding while suppressing noisy or low-confidence evidence.

\item We provide a comprehensive evaluation covering answer accuracy, retrieval quality, faithfulness, scalability, latency, and robustness. Results on \textsc{ScienceQA}, \textsc{MultiModalQA}, and \textsc{OK-VQA} show improved answer quality and evidence grounding over strong multimodal RAG, graph-retrieval, and vision--language baselines, with retrieval gains further demonstrated on \textsc{MultiModalQA}.

\end{itemize}

\section{Related Work}
\label{sec:related}

Multimodal QA spans vision--language modeling, structured knowledge integration, retrieval-augmented generation, and faithful answer generation. Instruction-following VLMs and multimodal LLMs, including BLIP-2~\cite{li2023blip}, LLaVA~\cite{liu2023llava}, InstructBLIP~\cite{dai2023instructblip}, MiniGPT-4~\cite{zhu2023minigpt4}, Flamingo~\cite{alayrac2022flamingo}, Qwen3-VL~\cite{bai2025qwenvl3}, and proprietary models such as GPT-5 and Gemini 2.5~\cite{openai2025gpt5,google2025gemini25}, have substantially improved image--text reasoning. However, their predictions often depend on dense visual tokens, long prompts, or implicit parametric knowledge, making evidence attribution difficult for multi-hop or knowledge-intensive questions. GraphLoom therefore focuses on explicit evidence retrieval and controlled use during decoding.

Multimodal knowledge graphs (MMKGs) provide a complementary route to structured reasoning. MMKGR~\cite{zheng2023mmkgr} performs multi-hop reasoning over MMKGs, while Lee et al.~\cite{lee2024multimodal} integrate MMKG representations with LLM-based multimodal reasoning. Other work studies MMKG construction from cross-modal evidence~\cite{liu2025aligning}. Pythia-RAG~\cite{ali2026pythiarag} constructs a unified MMKG from textual, visual, and commonsense evidence and retrieves query-conditioned subgraphs, whereas EvoGraph-R1~\cite{lin2026evograph} supports iterative retrieval, expansion, and graph refinement over evolving multimodal knowledge hypergraphs. GraphLoom differs by focusing on which retrieved graph memories should reach the decoder at each generation step.

Recent RAG methods increasingly emphasize structured, iterative, and utility-aware retrieval. G-Retriever, SKURG, and IRCoT combine graph retrieval, structured multimodal evidence, or intermediate reasoning with evidence acquisition~\cite{G-retriever,yang2023enhancing,trivedi2023ircot}, while CRAG adds retrieval-quality assessment and corrective retrieval~\cite{yan2024crag}. VisDoM~\cite{suri2024visdom} combines visual and textual retrieval for multimodal document QA, and Talk2Doc~\cite{khan2025talk2doc} uses weighted knowledge graphs within RAG for patient QA. More recent systems include HiKEY~\cite{shin2026hikey}, which performs hierarchical coarse-to-fine multimodal retrieval with token-efficient evidence subgraphs, and Luo et al.~\cite{luo2026utility}, who rank visual evidence by estimated downstream utility rather than semantic similarity alone. RARR~\cite{gao2023rarr} complements these approaches by retrieving supporting evidence and revising unsupported outputs. GraphLoom builds on these ideas through reliability-calibrated, decoder-time routing of graph memories.

Efficiency is also important in retrieval-augmented reasoning because long contexts, repeated retrieval, and growing KV caches increase latency and memory use. Attention-sink retention and sparse-attention methods reduce these costs~\cite{xiao2023streamingllm,xiao-etal-2025-efficient}, but are not designed for graph-evidence routing. GraphLoom combines bounded subgraph retrieval, sparse self-context, and decoder-time slot selection, while reporting both full-pipeline and cached online latency.

Faithfulness remains central to evidence-grounded QA. FActScore~\cite{min2023factscore} evaluates generations through atomic factual claims, while MiniCheck~\cite{tang2024minicheck} provides claim-level verification against grounding documents. Revision-based methods such as RARR~\cite{gao2023rarr} can further improve attribution and correct unsupported content. GraphLoom separates revision-time verification from final assessment through an independent automatic verifier and human evaluation.

Overall, GraphLoom differs from prior multimodal QA, MMKG, and RAG systems in three respects: it treats multimodal KG-RAG as controlled evidence management rather than simple retrieve-and-read; it performs reliability-calibrated graph-slot routing inside a frozen decoder; and it evaluates answer quality together with retrieval effectiveness, noisy-evidence scalability, threshold robustness, latency, and independently assessed faithfulness.

\begin{figure*}[t]
  \centering
  \includegraphics[width=15.5cm, height=14.0cm]{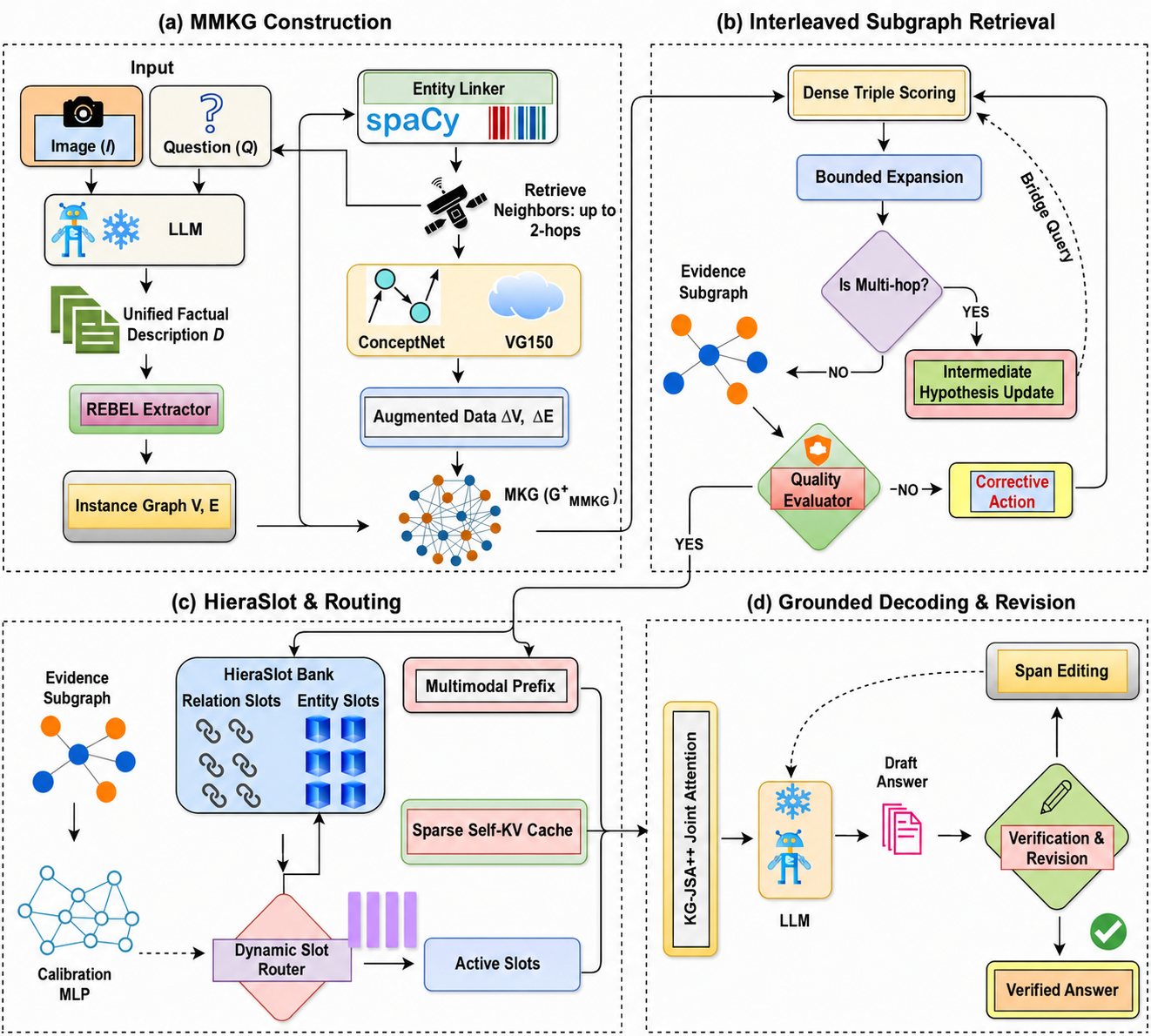}
  \caption{\textbf{GraphLoom} architecture. Given a question and its associated multimodal input, GraphLoom constructs an instance-level multimodal knowledge graph, retrieves a compact task-conditioned evidence subgraph, and organizes the retrieved evidence into multimodal prefix memories and hierarchical key--value graph slots. A reliability-calibrated router selects high-utility slots, which are injected into a frozen decoder through \textsc{KG-JSA++} joint graph--sequence attention. Interleaved retrieval and budgeted corrective retrieval improve multi-hop evidence coverage, while verification--revision supports faithful answer generation.}
  \label{fig:graphloom}
\end{figure*}

\section{Methodology: GraphLoom}
\label{sec:Model}

Figure~\ref{fig:graphloom} illustrates \textbf{GraphLoom}, a reliability-calibrated multimodal KG-RAG framework for controlled evidence retrieval, organization, and decoder-time injection. Given a question $Q$ and its associated multimodal input, including image $I$ when available, GraphLoom constructs an instance-level multimodal knowledge graph (MMKG) from grounded scene descriptions, extracted relational triples, and external visual/common-sense knowledge. It then retrieves a compact task-conditioned evidence subgraph using multimodal dense retrieval with bounded expansion. Retrieved triples are encoded as \textsc{HieraSlot} memories with a multimodal prefix. During decoding, a reliability-calibrated router selects Top-$k$ graph slots using combined utility--reliability scores, and \textsc{KG-JSA++} injects the selected slots and prefix memories into a frozen \textbf{Llama-3.1-8B-Instruct} decoder through joint graph--sequence attention.

GraphLoom separates \emph{evidence construction} from \emph{online evidence use}. For static multimodal collections, scene descriptions, triples, entity links, and enriched instance graphs can be precomputed and indexed. At query time, the online path includes query-conditioned retrieval, slot routing, decoder-time evidence injection, and conditional correction/revision. This enables the experiments to report both full end-to-end latency and cached online latency, clarifying where computational cost occurs.

For multi-hop questions, GraphLoom performs at most one additional hypothesis-guided retrieval round. If evidence is insufficient, a lightweight adequacy evaluator triggers at most one budgeted corrective action. After generation, a verification--revision stage edits unsupported spans using retrieved evidence. The revision-time verifier is treated as an internal component, while final faithfulness is evaluated with independent evaluators and human validation.

\subsection{Augmented Multimodal KG Construction}
\label{subsec:mmkg}

For each input instance, we construct an instance-level multimodal knowledge graph
$\mathcal{G}_{\mathrm{MMKG}}=(V,E)$ that represents extracted multimodal evidence under a shared $(s,r,o)$ schema. The graph construction is independent of the question $Q$, which is introduced later during task-conditioned retrieval. The graph is local to the input instance, but its triples are embedded in the same retrieval space as external candidate evidence, allowing the retrieval module to be evaluated under both compact local and larger noisy evidence-pool settings.

\textbf{Scene description and triple extraction.}
When an image $I$ is available, Qwen3-VL-Instruct~\cite{bai2025qwenvl3} generates a concise scene description $D$ grounded in the image. We then extract structured triples $t=(s,r,o)$ from $D$ using REBEL~\cite{rebel}. Each extracted triple is associated with a normalized extraction-confidence score
$\mathrm{conf}_{\mathrm{ext}}(t)\in[0,1]$, derived from the model's generation likelihood. Entity mentions are canonicalized using a lightweight named-entity and alias-linking step. Available textualized input evidence that is not represented as graph triples is retained separately for the evidence-bound revision context in Section~\ref{subsec:corrective}. For static image repositories, this construction step is cacheable and does not need to be repeated for every question over the same image.

\textbf{Confidence-aware filtering.}
To reduce extraction noise, we retain only triples satisfying
\begin{equation}
E=\left\{t \mid \mathrm{conf}_{\mathrm{ext}}(t)\geq
\tau_{\mathrm{edge}}\right\}.
\end{equation}
The default value is $\tau_{\mathrm{edge}}=0.5$. All filtering thresholds are selected using validation data only and then kept fixed during testing. We do not tune thresholds separately for individual test sets. The experimental section reports robustness under $\pm20\%$ perturbations of the main evidence-selection and verification thresholds.

\textbf{External KG enrichment.}
We enrich the instance graph with relevant neighbors from \textit{VG150} visual relations and \textit{ConceptNet} commonsense knowledge~\cite{speer2017conceptnet}. For each entity $e\in V$, we consider neighbors within at most two hops and retain relations whose normalized source score exceeds $\tau_{\mathrm{KB}}=0.7$, yielding
\begin{equation}
\mathcal{G}^{+}_{\mathrm{MMKG}}
=
\left(V\cup\Delta V,\; E\cup\Delta E\right).
\end{equation}
The enrichment is depth-bounded to two hops, while the subsequent retrieval stage applies explicit degree and total-edge budgets to prevent uncontrolled evidence expansion. In the standard setting, retrieval is performed over $\mathcal{G}^{+}_{\mathrm{MMKG}}$. In the scalability setting, $\mathcal{G}^{+}_{\mathrm{MMKG}}$ is mixed with an external distractor evidence pool to test retrieval robustness under noisy knowledge-management conditions.

\subsection{Compact Evidence Subgraph Retrieval}
\label{subsec:retrieval}

GraphLoom retrieves evidence from a candidate pool $\mathcal{C}(Q,I)$:
\begin{equation}
\mathcal{C}(Q,I)
=
\mathcal{E}^{+}_{\mathrm{MMKG}}
\cup
\mathcal{E}_{\mathrm{ext}},
\end{equation}
\noindent where $\mathcal{E}^{+}_{\mathrm{MMKG}}$ denotes triples from the enriched instance graph and $\mathcal{E}_{\mathrm{ext}}$ denotes optional external or distractor triples used in scalability experiments. In the standard instance-level setting, $\mathcal{E}_{\mathrm{ext}}=\emptyset$. This formulation keeps the core retrieval mechanism unchanged while allowing evaluation under larger and noisier candidate evidence pools.

Each candidate triple $t$ is verbalized as $\mathrm{text}(t)$ and scored by multimodal similarity:
\begin{equation}
\mathrm{score}(t,Q,I)
=
\cos\!\left(
f_{\mathrm{embed}}(I,Q),
f_{\mathrm{embed}}(\mathrm{text}(t))
\right),
\end{equation}
\noindent where $f_{\mathrm{embed}}$ denotes Qwen3-VL-Embedding~\cite{li2026qwen3vlembedding}; when no image is available, the query is embedded in text-only mode. For large candidate pools, triple embeddings are precomputed and indexed before query-time retrieval. This affects retrieval efficiency but does not change the scoring function.

\begin{equation}
\mathcal{G}_{\mathrm{sub}}(Q)
=
\left(
V_{\mathrm{sub}},
\mathcal{E}_{\mathrm{sub}}(Q)
\right),
\qquad
|\mathcal{E}_{\mathrm{sub}}(Q)| \leq N_{\max}.
\end{equation}

This bounded retrieval design prevents the decoder from receiving large flattened neighborhoods. The budgets $K$, $d_{\max}$, and $N_{\max}$ are selected on validation data and kept fixed during testing.

\subsection{Interleaved Multi-hop Retrieval}
\label{subsec:interleaved}

To recover missing connectors for multi-hop reasoning, GraphLoom performs at most two bounded retrieval rounds. From the initial subgraph $\mathcal{G}_{\mathrm{sub}}^{(1)}(Q)$, we select bridge nodes using degree and centrality signals. Candidate bridge nodes satisfy $\deg(v)\geq 2$ and are ranked by betweenness centrality within the retrieved subgraph.

A frozen \textbf{Llama-3.1-8B-Instruct} decoder then generates a concise intermediate query $\widetilde{Q}$ conditioned on $Q$ and the selected bridge evidence. To limit autoregressive overhead, $\widetilde{Q}$ is capped at 15 tokens and only one additional retrieval round is allowed. We rerun retrieval with $(I,\widetilde{Q})$ to obtain $\mathcal{G}_{\mathrm{sub}}^{(2)}(\widetilde{Q})$. The merged edge set is represented using following equation: 
\begin{equation}
\mathcal{E}_{\mathrm{mrg}}
=
\mathcal{E}_{\mathrm{sub}}^{(1)}(Q)
\cup
\mathcal{E}_{\mathrm{sub}}^{(2)}(\widetilde{Q}).
\end{equation}

From this merged set, we retain the top-$N_{\max}$ edges according to their original question-conditioned scores:
\begin{equation}
\mathcal{E}_{\mathrm{sub}}(Q)
=
\operatorname{TopN}_{\,t\in\mathcal{E}_{\mathrm{mrg}}}
\left(
\mathrm{score}(t,Q,I),
\min\!\left(N_{\max},|\mathcal{E}_{\mathrm{mrg}}|\right)
\right).
\end{equation}

This interleaved step improves evidence coverage for multi-hop questions while keeping the number of retrieval and generation calls bounded. Its cost is reported separately from initial retrieval and final decoding.

\subsection{Evidence Memories: Fusion and \textsc{HieraSlot}}
\label{subsec:memories}

Retrieved evidence conditions the frozen Llama-3.1-8B-Instruct decoder through two complementary memory channels: a global multimodal prefix and structured graph-slot memories.

\textbf{Global multimodal prefix.}
We concatenate the question $Q$ with a concise facts block $\mathcal{F}$ derived from $\mathcal{G}_{\mathrm{sub}}(Q)$, yielding $T=[Q;\mathcal{F}]$. Text tokens are encoded by the frozen Qwen3-VL-Embedding text encoder to obtain $H_{\text{text}}\in\mathbb{R}^{L_T\times d_{\text{emb}}}$, while visual tokens $H_{\text{vis}}\in\mathbb{R}^{L_V\times d_{\text{vis}}}$ are extracted by the frozen Qwen3-VL-Instruct vision encoder. We fuse the two modalities with a lightweight cross-attention module:
\begin{equation}
\label{eq:fusion}
H_{\text{fus}}
=
\mathrm{CrossAttn}(H_{\text{text}},H_{\text{vis}};\theta_{\text{fus}})
\in\mathbb{R}^{L_T\times d},
\end{equation}
\noindent where $\theta_{\text{fus}}$ are trainable. The fused representation is pooled to a fixed prefix length $L_p=32$ and converted into prefix key/value pairs $(K_{\text{fus}},V_{\text{fus}})$ for \textsc{KG-JSA++}.

\textbf{Structured slot memories (\textsc{HieraSlot}).}
Each retrieved triple $t=(s,r,o)\in\mathcal{E}_{\mathrm{sub}}(Q)$ is represented by two complementary slots: an entity slot with key $k_e=\mathrm{Emb}(s)$ and value $v_e=\mathrm{Emb}(r\oplus o)$, and a relation slot with key $k_r=\mathrm{Emb}(r)$ and value $v_r=\mathrm{Emb}(s\oplus o)$. The embedding projection is defined as:
\begin{equation}
\mathrm{Emb}(x)
=
W_{\mathrm{slot}} f_{\mathrm{embed}}^{\mathrm{text}}(x)
+
b_{\mathrm{slot}},
\end{equation}
which maps the text embedding into the decoder hidden space. For $M$ retrieved triples, the slot bank is given as follows: 
\begin{equation}
\label{eq:hieraslot-bank}
S_{\mathrm{HieraSlot}}
=
\left\{(k_e^{(i)},v_e^{(i)})\right\}_{i=1}^{M}
\cup
\left\{(k_r^{(i)},v_r^{(i)})\right\}_{i=1}^{M}.
\end{equation}
Entity slots capture entity-specific attributes, whereas relation slots encode relational patterns. During routing, these slot types compete independently, enabling fine-grained control over whether entity-level or relation-level evidence is used at a decoding step. The selected slot representations are subsequently mapped to the layer-specific key/value space used by \textsc{KG-JSA++}.

\subsection{Reliability-Calibrated Slot Routing}
\label{subsec:slot-routing}

At decoding step $t$, GraphLoom selects a small active set of graph slots from
$S_{\mathrm{HieraSlot}}$. For each candidate slot $j$, the router uses slot
content, retrieval score, extraction confidence, source type, graph centrality,
and the current decoder state to estimate a utility score $u_j$ and reliability
score $\hat{c}_j$:
\begin{equation}
u_j = \mathrm{MLP}_{u}(\phi_j), \qquad
\hat{c}_j = \sigma\!\left(\mathrm{MLP}_{c}(\phi_j)\right),
\end{equation}
\noindent where $\phi_j$ denotes the feature vector for slot $j$ and
$\sigma(\cdot)$ is the sigmoid function. Because the decoder-state component changes with decoding step $t$, $u_j$, $\hat{c}_j$, and $\rho_j$ are step-dependent; the index $t$ is suppressed for readability. The final routing score combines
semantic utility and reliability:
\begin{equation}
\rho_j
=
u_j
+
\beta\,\mathrm{logit}(\hat{c}_j),
\end{equation}
\noindent where $\beta$ controls the strength of reliability calibration.
The active slot set is defined as follows: 
\begin{equation}
\mathcal{A}_t
=
\operatorname{TopK}_{j\in S_{\mathrm{HieraSlot}}}
\left(\rho_j,k\right).
\end{equation}

Only slots in $\mathcal{A}_t$ are passed to \textsc{KG-JSA++}. This keeps
decoder-time evidence use compact and reduces exposure to noisy triples.
The active-slot budget $k$ is fixed after validation, and its sensitivity
is reported experimentally.

\subsection{\textsc{KG-JSA++}: Joint Attention with Sparse Self-KV}
\label{subsec:kgjsa}

\textsc{KG-JSA++} integrates external graph evidence with decoder self-attention through a single joint attention operation inside the frozen \textbf{Llama-3.1-8B-Instruct} decoder. Unlike separate cross-attention, it lets routed graph memories and internal context compete directly for attention, enabling fine-grained and reliability-aware evidence use.

At decoding layer $\ell$ and step $t$, GraphLoom forms prefixed memory by concatenating the fused multimodal prefix with the routed slot memories:
\begin{equation}
K_{\mathrm{pref}}^{(\ell)}(t)
=
\left[
K_{\mathrm{fus}}^{(\ell)};
\tilde{K}_{t}^{(\ell)}
\right],
\qquad
V_{\mathrm{pref}}^{(\ell)}(t)
=
\left[
V_{\mathrm{fus}}^{(\ell)};
\tilde{V}_{t}^{(\ell)}
\right],
\end{equation}
\noindent where $(\tilde{K}_{t}^{(\ell)},\tilde{V}_{t}^{(\ell)})$ are the key/value pairs corresponding to active slots in $\mathcal{A}_t$.

To improve decoding efficiency, we sparsify the self-attention key--value cache instead of attending to the full token history. Following the attention-sink principle of StreamingLLM~\cite{xiao2023streamingllm}, the retained cache includes: (i) recent tokens, (ii) sink tokens from the beginning of the sequence, and (iii) high-similarity prior tokens selected by cosine similarity to the current query $q_t^{(\ell)}$. Their union is defined as:
\begin{equation}
I_t^{(\ell)}
=
I_{\mathrm{recent}}
\cup
I_{\mathrm{sink}}
\cup
I_{\mathrm{sim}}.
\end{equation}

\textsc{KG-JSA++} computes attention logits over both external memories and sparse self-context:
\begin{equation}
\begin{aligned}
S_{\mathrm{pref}}
&=
\frac{
q_t^{(\ell)}
\left(K_{\mathrm{pref}}^{(\ell)}(t)\right)^\top
}{
\sqrt{d_h}
}
+
B_{\mathrm{pref}}(t),
\\
S_{\mathrm{self}}
&=
\frac{
q_t^{(\ell)}
\left(K_{\mathrm{self}}^{(\ell)}(t)
[I_t^{(\ell)}]\right)^\top
}{
\sqrt{d_h}
}
+
M_{\mathrm{causal}}[I_t^{(\ell)}],
\end{aligned}
\label{eq:kgjsa}
\end{equation}
\noindent where $d_h$ is the attention-head dimension and $M_{\mathrm{causal}}$ enforces causal masking. To encourage reliable evidence without overriding semantic relevance, slot positions receive a reliability-aware logit bias:
\begin{equation}
B_{\mathrm{pref}}(t)[j]
=
\begin{cases}
\gamma\,\mathrm{logit}(\hat{c}_j),
& j \in \text{slot positions},\\
0,
& j \in \text{fused-prefix positions},
\end{cases}
\end{equation}
\noindent where $\gamma$ is fixed after validation.

A single softmax is applied over the concatenated prefix and self-attention scores:
\begin{equation}
\alpha_t^{(\ell)}
=
\mathrm{softmax}
\left(
\left[
S_{\mathrm{pref}};
S_{\mathrm{self}}
\right]
\right).
\end{equation}
The output is computed as follows:
\begin{equation}
o_t^{(\ell)}
=
\alpha_t^{(\ell)}
\begin{bmatrix}
V_{\mathrm{pref}}^{(\ell)}(t)\\
V_{\mathrm{self}}^{(\ell)}(t)[I_t^{(\ell)}]
\end{bmatrix}.
\end{equation}

For fixed head dimension and a fixed retained sparse context, the per-step attention-score computation scales as:
\begin{equation}
\mathcal{O}
\left(
L_p
+
|\mathcal{A}_t|
+
|I_t^{(\ell)}|
\right),
\end{equation}
compared with $\mathcal{O}(t)$ for full causal self-attention. Thus, the attention cost depends on the fixed prefix length, active slot budget, and retained sparse self-context size rather than directly on the full generated history.

\subsection{Corrective Retrieval and Evidence-Bound Revision}
\label{subsec:corrective}

To mitigate failures from imperfect retrieval, GraphLoom uses a lightweight evidence-adequacy evaluator that can trigger budgeted corrective retrieval, followed by a verification--revision pass over the generated answer. The evaluator computes:
\begin{equation}
E_{\mathrm{eval}}
=
\sigma\!\left(
\mathrm{MLP}\bigl(
g_{\mathrm{graph}}
\oplus
g_{\mathrm{query}}
\oplus
g_{\mathrm{history}}
\bigr)
\right),
\end{equation}
\noindent where $E_{\mathrm{eval}}\in(0,1)$, $g_{\mathrm{graph}}$ summarizes subgraph statistics, $g_{\mathrm{query}}$ captures question difficulty, and $g_{\mathrm{history}}$ encodes prior retrieval actions. If $E_{\mathrm{eval}}<\tau_{\mathrm{eval}}$, GraphLoom executes at most one corrective action.

We consider three bounded actions: (i) \emph{Expansion Boost}, which temporarily increases $N_{\max}$ by a fixed validation-selected ratio and reruns bounded expansion; (ii) \emph{Query Reformulation}, which generates a concise clarified query conditioned on $Q$ and $\mathcal{G}_{\mathrm{sub}}(Q)$; and (iii) \emph{Verbalization Repair}, which generates alternative verbalizations for low-confidence triples and re-scores them with the multimodal embedder. The action selector trades off expected utility against the remaining evidence budget. No iterative correction loop is used.

After answer generation, a verification--revision stage enforces evidence faithfulness. The draft answer $A_{\mathrm{draft}}$ is decomposed into atomic claims $\{c_i\}$. Each claim is checked using a DeBERTa-v3 NLI verifier~\cite{he2021debertav3} against a textualized evidence context composed of the retrieved subgraph, grounded scene description, and available textual evidence:
\begin{equation}
\mathcal{E}_{\mathrm{rev}}
=
\mathrm{Textualize}\!\left(\mathcal{G}_{\mathrm{sub}}(Q)\right)
\oplus D
\oplus C_{\mathrm{text}},
\quad
e_i
=
\mathrm{NLI}_{\mathrm{rev}}
\left(c_i,\mathcal{E}_{\mathrm{rev}}\right),
\end{equation}
\noindent where $C_{\mathrm{text}}$ denotes available textualized input evidence, including serialized table context when applicable. Claims with $e_i<\tau_{\mathrm{ent}}$ trigger retrieval of the top relevant triples from $\mathcal{G}^{+}_{\mathrm{MMKG}}$, followed by a constrained revision prompt to the frozen decoder. The prompt allows edits only to unsupported spans and preserves supported spans whenever possible.

Importantly, the revision-time verifier is not used as the sole final faithfulness evaluator. The experimental section reports final faithfulness using independent verification and human validation, preventing the H-Rate results from being tied only to the same NLI model used inside revision.

\subsection{Answer Prediction}
\label{subsec:answer}

\textbf{Multiple-choice QA.}
Given answer options $\{o^{(j)}\}_{j=1}^{n}$, each option is scored by its length-normalized log-likelihood under the frozen decoder conditioned on the memory context $\mathcal{M}$:
\begin{equation}
\label{eq:mcq-score}
S(o^{(j)})
=
\frac{1}{|o^{(j)}|}
\sum_{i=1}^{|o^{(j)}|}
\log P\!\left(
y_i^{(j)}
\mid
y_{<i}^{(j)},\mathcal{M}
\right).
\end{equation}
The predicted answer is computed as:
\begin{equation}
\hat{y}
=
\arg\max_j S(o^{(j)}).
\end{equation}

\textbf{Open-ended QA.}
For free-form generation, we use autoregressive decoding conditioned on $\mathcal{M}$. To keep evaluation reproducible, decoding settings are fixed across all controlled comparisons. If sampling is used in any setting, the random seed and number of runs are fixed consistently.

\noindent\textbf{Evidence insufficiency.}
GraphLoom includes an evidence-safety fallback that can emit a special
\texttt{[NO\_EVIDENCE]} token when the retrieved evidence does not satisfy the adequacy requirement. In this case, decoding stops and the system returns ``Insufficient evidence.'' To avoid treating abstention as a source of artificial performance gain, such outputs are counted as incorrect for answer correctness, while their occurrence is tracked separately from claim-level faithfulness evaluation.

\subsection{Computation and Latency Accounting}
\label{subsec:latency-accounting}

GraphLoom contains both cacheable graph-construction and query-time inference stages. We therefore report three complementary costs. \textbf{Full-pipeline latency} includes scene description generation, triple extraction, entity linking, KG enrichment, retrieval, optional interleaved/corrective retrieval, decoding, verification, and revision. \textbf{Online latency} assumes cached instance-level graph construction and includes only query-conditioned retrieval, slot routing, decoding, and conditional correction/revision. \textbf{Decoder-time cost} reports the active slot count $|\mathcal{A}_t|$, prefix length $L_p$, sparse self-context size $|I_t^{(\ell)}|$, and the resulting per-step attention cost. This separation distinguishes cacheable preprocessing from the cost introduced during online evidence routing and generation.

\subsection{Training Objective}
\label{sec:training_objective}

All backbone models, including the Qwen3-VL encoders and the frozen Llama-3.1-8B-Instruct decoder, remain fixed during training. The main end-to-end training objective optimizes the lightweight multimodal fusion projection, \textsc{HieraSlot} projections, and reliability router. It combines answer generation, router calibration, contrastive slot utility, and router--attention alignment:
\begin{equation}
\mathcal{L}
=
\lambda_{\mathrm{gen}}\mathcal{L}_{\mathrm{gen}}
+
\lambda_{\mathrm{cal}}\mathcal{L}_{\mathrm{cal}}
+
\lambda_{\mathrm{cf}}\mathcal{L}_{\mathrm{cf}}
+
\lambda_{\mathrm{align}}\mathcal{L}_{\mathrm{align}}.
\end{equation}
The validation-selected weights are
$\{\lambda_{\mathrm{gen}},\lambda_{\mathrm{cal}},
\lambda_{\mathrm{cf}},\lambda_{\mathrm{align}}\}
=
\{1.0,0.5,0.3,0.2\}$ and are fixed for test evaluation.

$\mathcal{L}_{\mathrm{gen}}$ is a label-smoothed negative log-likelihood loss over the gold answer sequence conditioned on the memory context $\mathcal{M}$ produced by the multimodal prefix and routed graph slots. For multiple-choice QA, it is applied to the gold option; for open-ended QA, it is applied to the reference answer.

$\mathcal{L}_{\mathrm{cal}}$ calibrates router confidence $\hat{c}_j$ using binary cross-entropy against a binary utility target $z_j$, indicating whether a slot provides supporting evidence for the target answer.
To further separate useful and misleading slots, $\mathcal{L}_{\mathrm{cf}}$ applies a margin loss to positive--negative slot pairs:
\begin{equation}
\mathcal{L}_{\mathrm{cf}}
=
\frac{1}{|\mathcal{P}|}
\sum_{(j^{+},j^{-})\in\mathcal{P}}
\max\!\left(
0,\,
m-\rho_{j^{+}}+\rho_{j^{-}}
\right),
\end{equation}
\noindent where $m$ is a validation-selected margin and $\mathcal{P}$ contains useful--misleading slot pairs.

Finally, $\mathcal{L}_{\mathrm{align}}$ aligns the router distribution with the decoder attention assigned to the selected evidence slots:
\begin{equation}
\mathcal{L}_{\mathrm{align}}
=
D_{\mathrm{KL}}
\left(
p_{\mathrm{route}}(\cdot)
\,\|\, 
p_{\mathrm{attn}}(\cdot)
\right).
\end{equation}
Here, $p_{\mathrm{attn}}$ is obtained by averaging attention to selected slots across decoder heads, layers, and answer-token steps and renormalizing over the same slot set as $p_{\mathrm{route}}$. This encourages selected evidence to align with evidence actually used during generation without using the final faithfulness evaluator as a training signal.

The trainable modules are optimized with AdamW, linear warmup, cosine decay, and gradient clipping at 1.0. Early stopping uses
$S_{\mathrm{val}}=0.7S_{\mathrm{ans}}+0.3S_{\mathrm{faith}}$,
where $S_{\mathrm{ans}}$ is the validation answer score and $S_{\mathrm{faith}}$ is a validation-only grounding score. All loss weights, thresholds, and routing/retrieval budgets are selected on validation data and fixed before test evaluation.

\section{Experimental Evaluation}
\label{sec:Evaluation}

We evaluate \textbf{GraphLoom} along six axes: (i) answer correctness, (ii) retrieval quality under strict budgets, (iii) independently assessed faithfulness, (iv) scalability under noisy candidate pools, (v) robustness to threshold variation, and (vi) efficiency with separate full-pipeline and online latency accounting.

\subsection{Datasets and Evaluation Protocol}
\label{subsec:datasets_protocol}

\textbf{Datasets.}
We evaluate on three complementary multimodal QA benchmarks. \textbf{ScienceQA}~\cite{Lu2022ScienceQA} is a multiple-choice benchmark with image and context inputs (4{,}241 test instances) and no gold supporting evidence. \textbf{OK-VQA}~\cite{marino2019okvqa} is an open-ended benchmark requiring external knowledge, with 5{,}046 test instances. \textbf{MultiModalQA (MMQA)}~\cite{talmor2021multimodalqa} requires reasoning across text, tables, and images. Because answers and supporting-context annotations for the public MMQA test split are unavailable, we report MMQA answer and retrieval results on the official labeled development split (2{,}441 instances). Standard benchmark splits are otherwise retained.

\textbf{Answer correctness.}
We report \emph{Accuracy} on ScienceQA, standard \emph{VQA accuracy} on OK-VQA, and \emph{Exact Match (EM)} and token-level \emph{F1} on MMQA using the corresponding evaluation scripts.

\textbf{Retrieval quality.}
On the labeled MMQA split, we evaluate the retrieved subgraph $\mathcal{G}_{\mathrm{sub}}$ using \emph{Precision@5}, \emph{Recall@5}, and \emph{MRR} against gold supporting evidence. Semantic alignment between retrieved triples and gold evidence is measured with an evaluator not used during answer revision. We also evaluate retrieval under larger candidate pools containing distractor triples, as described in Section~\ref{subsec:scalability_eval}.

\textbf{Faithfulness.}
We assess grounding using \emph{Hallucination Rate (H-Rate)} following atomic-fact verification~\cite{min2023factscore}. Generated answers are decomposed into atomic claims and checked against retrieved and input evidence. Because GraphLoom uses an NLI verifier during revision, final H-Rate is evaluated with an independent automatic verifier and complementary human validation. Lower H-Rate indicates fewer unsupported claims.

\textbf{Efficiency.}
We report \emph{decoder-time cost} through the retrieved-edge count $|\mathcal{E}_{\mathrm{sub}}|$, active \textsc{HieraSlot} count $|\mathcal{A}_t|$, sparse self-KV size $|I_t^{(\ell)}|$, and per-step attention-FLOP reduction. \emph{Online latency} assumes cached graph construction and includes query-conditioned retrieval, slot routing, decoding, and conditional correction/revision. \emph{Full-pipeline latency} additionally includes scene description generation, triple extraction, entity linking, and KG enrichment.

\subsection{Baselines}
\label{subsec:baselines}

We compare GraphLoom against retrieval-free VLMs, multimodal RAG/KG-QA systems, retrieval-control and revision frameworks, and targeted ablations.

\textbf{Vision--Language Models (VLMs).}
We include LLaVA-1.5~\cite{liu2023llava} as a widely used open-source baseline, together with recent models Qwen2.5-VL~\cite{bai2025qwen25vl}, InternVL3~\cite{zhu2025internvl3}, and LLaVA-OneVision-1.5~\cite{an2025llavaonevision15}. Qwen3-VL-Instruct~\cite{bai2025qwenvl3}, which is also used in GraphLoom's graph-construction pipeline, is evaluated as an additional open-source baseline. We further include \textbf{GPT-5}~\cite{openai2025gpt5} and \textbf{Gemini~2.5~Pro}~\cite{google2025gemini25} as closed-source reference models; because their APIs may change over time, they are not treated as controlled baselines.

\textbf{Multimodal RAG/KG-QA systems.}
We compare with MuRAG~\cite{chen2022murag} and SKURG~\cite{yang2023enhancing} to assess the benefit of instance-level MMKG construction and compact slot-based evidence injection over document-level or prebuilt-graph retrieval.

\textbf{Retrieval-control / revision frameworks.}
We re-implement IRCoT-style interleaved retrieval and the corrective trigger from CRAG~\cite{trivedi2023ircot,yan2024crag} using the same retriever and decoder backbone. These variants use Qwen3-VL-Embedding for retrieval and Llama-3.1-8B-Instruct for decoding, but exclude the reliability-calibrated router and \textsc{KG-JSA++}. For the faithfulness comparison, we additionally include RARR~\cite{gao2023rarr} as a revision-based baseline. This isolates the contribution of GraphLoom's controlled evidence-routing and revision mechanisms.

\textbf{Ablations.}
\textbf{Flat-RAG} injects retrieved evidence as plain text using the same retriever and evidence sources, measuring the effect of structured slot memories. \textbf{CLIP-RAG} replaces Qwen3-VL-Embedding with CLIP-ViT-L/14~\cite{radford2021learning} while keeping the remaining pipeline unchanged, isolating gains beyond retriever choice. Additional ablations remove interleaved retrieval, corrective retrieval, post-hoc revision, reliability-calibrated routing, and \textsc{KG-JSA++}.

\subsection{Overall Results}
\label{subsec:overall_results}

Tables~\ref{tab:overall_qa_summary} and~\ref{tab:okvqa_results} summarize QA performance on ScienceQA, MMQA, and OK-VQA. All controlled open-source methods and GraphLoom variants use the same evaluation splits, prompts, and protocol, with mean $\pm$ std reported over five runs. Closed-source API models are reported only as reference comparisons.

\paragraph{Comparative Analysis.}
GraphLoom achieves 92.5\% accuracy on ScienceQA, 55.2 EM / 66.5 F1 on MMQA, and 70.5\% VQA accuracy on OK-VQA. The gain is modest on ScienceQA, where recent VLMs already perform strongly, but larger on MMQA and OK-VQA, which require external knowledge, multimodal evidence alignment, or multi-hop reasoning. These results support the benefit of explicit graph retrieval and controlled evidence injection for knowledge-intensive multimodal QA.

\begin{table}[!t]
\centering
\caption{QA results on ScienceQA and MMQA. Bold indicates the best controlled open-source result; closed-source models are reference-only comparisons.}
\label{tab:overall_qa_summary}
\scalebox{0.74}{
\begin{tabular}{|l|c|c|c|}\hline
\rowcolor{gray!15}
\textbf{Model} 
& \textbf{ScienceQA Acc} $\uparrow$
& \textbf{MMQA EM} $\uparrow$
& \textbf{MMQA F1} $\uparrow$ \\\hline
LLaVA-1.5 (7B)~\cite{liu2023llava} & 89.3 $\pm$ 0.2 & 44.8 $\pm$ 0.3 & 54.9 $\pm$ 0.4 \\
Qwen2.5-VL~\cite{bai2025qwen25vl} & 91.2 $\pm$ 0.2 & 50.6 $\pm$ 0.4 & 61.9 $\pm$ 0.4 \\
InternVL3~\cite{zhu2025internvl3} & 91.8 $\pm$ 0.2 & 51.4 $\pm$ 0.4 & 62.8 $\pm$ 0.4 \\
LLaVA-OneVision-1.5~\cite{an2025llavaonevision15} & 91.5 $\pm$ 0.3 & 50.9 $\pm$ 0.5 & 62.3 $\pm$ 0.4 \\
Qwen3-VL-Instruct (2B)~\cite{bai2025qwenvl3} & 89.5 $\pm$ 0.2 & 46.4 $\pm$ 0.4 & 55.9 $\pm$ 0.3 \\
MuRAG~\cite{chen2022murag} & 87.9 $\pm$ 0.3 & 47.8 $\pm$ 0.5 & 57.8 $\pm$ 0.4 \\
SKURG~\cite{yang2023enhancing} & 88.4 $\pm$ 0.3 & 48.3 $\pm$ 0.4 & 58.3 $\pm$ 0.5 \\
IRCoT~\cite{trivedi2023ircot} & 89.7 $\pm$ 0.2 & 49.3 $\pm$ 0.4 & 59.3 $\pm$ 0.3 \\
CRAG~\cite{yan2024crag} & 89.4 $\pm$ 0.3 & 48.9 $\pm$ 0.5 & 58.9 $\pm$ 0.4 \\\hline
\multicolumn{4}{|c|}{\textbf{Closed-Source Reference Models}} \\\hline
GPT-5~\cite{openai2025gpt5} & 92.8 & 54.2 & 65.7 \\
Gemini 2.5 Pro~\cite{google2025gemini25} & 93.1 & 54.8 & 66.3 \\\hline
\multicolumn{4}{|c|}{\textbf{GraphLoom Ablations}} \\\hline
Flat-RAG & 90.1 $\pm$ 0.3 & 49.8 $\pm$ 0.4 & 61.2 $\pm$ 0.4 \\
CLIP-RAG & 89.8 $\pm$ 0.2 & 49.0 $\pm$ 0.5 & 60.5 $\pm$ 0.5 \\\hline
\textbf{GraphLoom} & \textbf{92.5 $\pm$ 0.2} & \textbf{55.2 $\pm$ 0.4} & \textbf{66.5 $\pm$ 0.3} \\\hline
\end{tabular}
}
\end{table}

\begin{table}[!t]
\centering
\caption{Results on OK-VQA. GraphLoom achieves the strongest controlled open-source score and remains competitive with closed-source reference models.}
\label{tab:okvqa_results}
\scalebox{0.92}{
\begin{tabular}{|l|c|}\hline
\rowcolor{gray!15}
\textbf{Model}
& \textbf{VQA Accuracy} $\uparrow$ \\\hline
LLaVA-1.5 (7B)~\cite{liu2023llava} & 61.2 $\pm$ 0.4 \\
Qwen2.5-VL~\cite{bai2025qwen25vl} & 66.2 $\pm$ 0.4 \\
InternVL3~\cite{zhu2025internvl3} & 67.4 $\pm$ 0.3 \\
LLaVA-OneVision-1.5~\cite{an2025llavaonevision15} & 66.8 $\pm$ 0.4 \\
Qwen3-VL-Instruct (2B)~\cite{bai2025qwenvl3} & 63.8 $\pm$ 0.3 \\
MuRAG~\cite{chen2022murag} & 59.5 $\pm$ 0.5 \\
SKURG~\cite{yang2023enhancing} & 62.1 $\pm$ 0.4 \\
IRCoT~\cite{trivedi2023ircot} & 64.3 $\pm$ 0.4 \\
CRAG~\cite{yan2024crag} & 63.9 $\pm$ 0.5 \\\hline
GPT-5~\cite{openai2025gpt5} & 68.7 \\
Gemini 2.5 Pro~\cite{google2025gemini25} & 69.5 \\\hline
\textbf{GraphLoom} & \textbf{70.5 $\pm$ 0.3} \\\hline
\end{tabular}
}
\end{table}

Compared with IRCoT and CRAG, GraphLoom achieves higher MMQA scores, suggesting that reliability-calibrated routing and \textsc{KG-JSA++} provide benefits beyond interleaved or corrective retrieval alone. Flat-RAG and CLIP-RAG also perform below GraphLoom, indicating that the gains are associated not only with retrieving evidence but also with organizing it into structured slots and selectively routing graph memories during decoding. GraphLoom also remains competitive with GPT-5 and Gemini~2.5~Pro under our evaluation protocol, although these closed-source systems are treated only as reference comparisons.

\paragraph{Faithfulness.}
Detailed faithfulness results are reported in Section~\ref{subsec:hallucination_eval} using independent automatic H-Rate and complementary human validation. GraphLoom achieves the lowest unsupported-claim rate among the evaluated systems, consistent with the intended roles of reliability-calibrated routing and evidence-bound revision.

\subsection{Ablation Studies}
\label{subsec:ablations}

We ablate the main GraphLoom components on MMQA to determine whether the gains arise from retrieval alone or from the proposed evidence-routing design. The ablations remove interleaved retrieval, corrective retrieval, post-hoc revision, reliability-calibrated routing, and \textsc{KG-JSA++}; we also analyze routing features, external knowledge sources, and router calibration.

Table~\ref{tab:ablations} shows that all components contribute to performance, with the largest drops occurring when the reliability-calibrated router or \textsc{KG-JSA++} is removed. Without the router, H-Rate increases from 10.5 to 17.3, consistent with reduced suppression of low-confidence or weakly relevant evidence. Removing \textsc{KG-JSA++} causes the largest degradation in both answer quality and faithfulness, indicating the importance of jointly attending to routed graph memories and decoder context. Overall, the results suggest that GraphLoom benefits not only from retrieving evidence, but also from controlling how that evidence is selected and injected during decoding.

\begin{table}[!t]
\centering
\caption{Ablation on MMQA. Removing key components lowers EM/F1 and increases H-Rate, highlighting the importance of controlled evidence routing and joint graph--sequence attention.}
\label{tab:ablations}
\scalebox{0.85}{
\begin{tabular}{|l|c|c|c|}\hline
      \rowcolor{gray!15}
      \textbf{Variant} 
    & \textbf{EM} $\uparrow$ 
    & \textbf{F1} $\uparrow$ 
    & \textbf{H-Rate} $\downarrow$ \\\hline
        \textbf{Full \textsc{GraphLoom}} & \textbf{55.2 $\pm$ 0.4} & \textbf{66.5 $\pm$ 0.3} & \textbf{10.5 $\pm$ 0.4} \\\hline
        w/o Interleaved Retrieval & 53.6 $\pm$ 0.5 & 64.6 $\pm$ 0.4 & 11.1 $\pm$ 0.5 \\
        w/o Corrective Retrieval & 54.1 $\pm$ 0.4 & 65.0 $\pm$ 0.5 & 11.8 $\pm$ 0.6 \\
        w/o Post-hoc Revision & 54.7 $\pm$ 0.4 & 65.7 $\pm$ 0.4 & 13.9 $\pm$ 0.5 \\
        w/o Reliability-Calibrated Router & 51.8 $\pm$ 0.6 & 61.6 $\pm$ 0.5 & 17.3 $\pm$ 0.7 \\
        w/o KG-JSA++ & 50.5 $\pm$ 0.7 & 60.3 $\pm$ 0.6 & 18.9 $\pm$ 0.8 \\\hline
\end{tabular}
}
\end{table}

Interleaved and corrective retrieval provide smaller but consistent gains. Removing interleaved retrieval mainly reduces EM/F1, consistent with the additional hypothesis-guided retrieval round recovering useful multi-hop connectors. Corrective retrieval has a larger effect on H-Rate, while post-hoc revision produces only modest EM/F1 changes but substantially reduces unsupported claims, consistent with its role as an evidence-bound correction step.

\textbf{Fine-grained ablations.}
Removing the retrieval score (\textit{-ret-score}) produces the largest single-feature degradation (EM $-1.6$, H-Rate $+2.8$), indicating that query--evidence alignment is a particularly important routing signal. Removing contextual features (\textit{-context}) also reduces performance (EM $-1.2$, H-Rate $+2.1$), while extractor confidence (\textit{-confext}) and centrality (\textit{-centrality}) have smaller but consistent effects. These results support combining semantic relevance, evidence reliability, and graph-structural cues rather than relying on a single signal.

\textbf{Knowledge sources.}
VG150 and ConceptNet provide complementary external evidence. Removing VG150 increases H-Rate by 0.8 points, consistent with its contribution to visual-relation grounding. Removing ConceptNet reduces EM by 1.3 points, with larger effects on questions requiring commonsense links beyond directly visible content. Removing both sources (\textit{No-KG}) yields the largest degradation, reducing EM by 2.4 points and increasing H-Rate by 4.8 points. This indicates that external knowledge contributes to performance when combined with bounded retrieval and reliability-aware routing.

\begin{figure}[t]
\centering
\begin{tikzpicture}
\begin{axis}[
        width=0.98\linewidth,
        height=5.2cm,
        xmin=0, xmax=1,
        ymin=0, ymax=1,
        xlabel={Predicted confidence},
        ylabel={Observed slot utility},
        xtick={0,0.2,0.4,0.6,0.8,1.0},
        ytick={0,0.2,0.4,0.6,0.8,1.0},
        grid=both,
        grid style={line width=.08pt, draw=gray!15},
        major grid style={line width=.12pt, draw=gray!25},
        ticklabel style={font=\scriptsize},
        label style={font=\scriptsize},
        legend style={
            at={(0.03,0.97)},
            anchor=north west,
            font=\scriptsize,
            draw=none,
            fill=white,
            fill opacity=0.85,
            text opacity=1
        },
        legend columns=1
        ]

\addplot[black, dashed, thick] coordinates {(0,0) (1,1)};
\addlegendentry{Perfect calibration}

\addplot[mark=o, mark size=1.5pt, thick, color=blue]
coordinates {
(0.05,0.06) (0.15,0.12) (0.25,0.27) (0.35,0.35) (0.45,0.44)
(0.55,0.56) (0.65,0.66) (0.75,0.78) (0.85,0.86) (0.95,0.97)
};
\addlegendentry{Calibrated router, ECE = 0.032}

\addplot[mark=*, mark size=1.5pt, thick, dashed, color=orange]
coordinates {
(0.05,0.01) (0.15,0.07) (0.25,0.32) (0.35,0.23) (0.45,0.54)
(0.55,0.63) (0.65,0.57) (0.75,0.88) (0.85,0.79) (0.95,0.98)
};
\addlegendentry{Uncalibrated router, ECE = 0.087}

\node[
    anchor=south east,
    fill=white,
    fill opacity=0.85,
    text opacity=1,
    draw=gray!35,
    rounded corners=1pt,
    font=\scriptsize,
    align=left
] at (axis cs:0.98,0.05)
{Lower ECE indicates\\more reliable slot selection};

\end{axis}
\end{tikzpicture}
\caption{Reliability diagram for slot routing. The calibrated router lies closer to the perfect-calibration diagonal, reducing ECE from 0.087 to 0.032.}
\label{fig:calibration}
\end{figure}
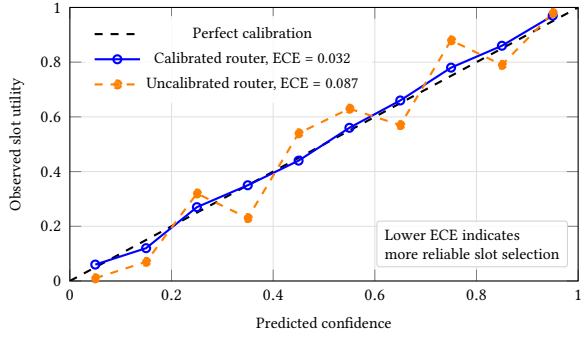

\textbf{Router calibration.}
Figure~\ref{fig:calibration} shows that the calibrated router is better aligned with empirical slot utility than the uncalibrated variant, reducing ECE from 0.087 to 0.032. This supports the reliability-calibration objective and is consistent with the improved faithfulness observed in Table~\ref{tab:ablations}.

\subsection{Scalability under Noisy Evidence Pools}
\label{subsec:scalability_eval}

To evaluate GraphLoom beyond compact instance-level graphs, we test retrieval under progressively larger candidate pools. For each MMQA instance, the enriched instance graph is mixed with distractor triples sampled from other MMQA instances and external KG triples unrelated to the current question. Triple embeddings are precomputed and indexed, so only the candidate-pool size changes; the scoring function and bounded-expansion mechanism remain fixed.

\begin{table}[!h]
\centering
\caption{Scalability under noisy candidate evidence pools on MMQA. GraphLoom degrades gradually as the retrieval space grows to one million distractor triples.}
\label{tab:scalability}
\scalebox{0.75}{
\begin{tabular}{|l|c|c|c|c|}\hline
      \rowcolor{gray!15}
      \textbf{Candidate pool} 
    & \textbf{Recall@5} $\uparrow$ 
    & \textbf{MRR} $\uparrow$ 
    & \textbf{MMQA F1} $\uparrow$ 
    & \textbf{Online latency (ms)} $\downarrow$ \\\hline
        Instance graph only       & 78.4 & 0.742 & 66.5 & 119 \\
        +10K distractor triples   & 77.6 & 0.734 & 66.2 & 124 \\
        +100K distractor triples  & 75.9 & 0.713 & 65.8 & 132 \\
        +1M distractor triples    & 73.2 & 0.684 & 65.1 & 148 \\\hline
\end{tabular}}
\end{table}

Table~\ref{tab:scalability} shows gradual degradation as the candidate pool grows. Moving from the instance graph to one million distractor triples reduces Recall@5 by 5.2 points and MRR by 0.058, while MMQA F1 drops by only 1.4 points. This suggests that bounded expansion and reliability-aware slot selection help limit the effect of noisy candidates on answer generation.

Online latency increases from 119\,ms to 148\,ms as retrieval operates over a larger indexed pool. However, the decoder-time budget remains fixed through $N_{\max}$ and the active-slot limit $k$, so generation cost does not scale with candidate-pool size. These results indicate that GraphLoom remains effective under substantially larger noisy retrieval spaces.

\subsection{Threshold Robustness}
\label{subsec:threshold_robustness}

To assess sensitivity to threshold selection, we perturb the main evidence-selection and verification thresholds by $\pm20\%$ around their validation-selected values while keeping all other settings fixed. This evaluates the stability of confidence filtering, KG enrichment, corrective retrieval, and evidence verification.

\begin{table}[!h]
\centering
\caption{Robustness under $\pm20\%$ threshold perturbations on MMQA. Performance remains stable under moderate changes to the main evidence-selection and verification thresholds.}
\label{tab:threshold_robustness}
\scalebox{0.88}{
\begin{tabular}{|l|c|c|c|}\hline
      \rowcolor{gray!15}
      \textbf{Setting} 
    & \textbf{MMQA EM} $\uparrow$ 
    & \textbf{MMQA F1} $\uparrow$ 
    & \textbf{H-Rate} $\downarrow$ \\\hline
        Default & \textbf{55.2} & \textbf{66.5} & 10.5 \\
        $\tau_{\mathrm{edge}}-20\%$ & 54.8 & 66.1 & 11.2 \\
        $\tau_{\mathrm{edge}}+20\%$ & 54.5 & 65.8 & 10.9 \\
        $\tau_{\mathrm{KB}}-20\%$ & 54.7 & 66.0 & 11.3 \\
        $\tau_{\mathrm{KB}}+20\%$ & 54.4 & 65.7 & 11.0 \\
        $\tau_{\mathrm{eval}}-20\%$ & 54.6 & 65.9 & 11.8 \\
        $\tau_{\mathrm{eval}}+20\%$ & 54.9 & 66.2 & 10.8 \\
        $\tau_{\mathrm{ent}}-20\%$ & 54.9 & 66.0 & 12.2 \\
        $\tau_{\mathrm{ent}}+20\%$ & 54.6 & 65.7 & \textbf{10.3} \\\hline
\end{tabular}
}
\end{table}

Table~\ref{tab:threshold_robustness} shows that MMQA EM and F1 remain within 0.8 points of the default setting across all perturbations, indicating limited sensitivity to moderate threshold changes.

Lowering $\tau_{\mathrm{edge}}$ or $\tau_{\mathrm{KB}}$ admits more evidence but also increases H-Rate slightly, whereas higher values reduce noise at the cost of some useful evidence. Reducing $\tau_{\mathrm{eval}}$ triggers fewer corrective actions and increases H-Rate, while increasing it produces a modest faithfulness improvement. Similarly, a higher $\tau_{\mathrm{ent}}$ makes revision stricter, lowering H-Rate but slightly reducing answer quality. Overall, the default setting provides the best accuracy--faithfulness balance.

\subsection{Quantitative Hallucination Assessment}
\label{subsec:hallucination_eval}

We assess unsupported assertions using complementary automatic and human evaluation, with the same prompts, decoding settings, and claim-splitting procedure across models.

\textbf{Independent automatic H-Rate.}
H-Rate is the fraction of atomic factual claims unsupported by the retrieved subgraph $\mathcal{G}_{\mathrm{sub}}$ or textualized input evidence. Because GraphLoom uses DeBERTa-v3 only during verification--revision, final H-Rate is computed with \textsc{MiniCheck}~\cite{tang2024minicheck}, which is not used during training or revision. Each claim is paired with an evidence context containing verbalized triples, retrieved facts, and the grounded scene description. We evaluate 1{,}500 examples, with 500 sampled from each dataset, and report mean $\pm$ std across runs.

\textbf{Human validation.}
We additionally evaluate 300 randomly sampled outputs, with 100 examples from each dataset. Three NLP researchers independently assess each output using an evidence-grounding rubric and majority voting. Visual claims are considered supported only when the corresponding visual or retrieved evidence is explicit. Thus, the human evaluation complements MiniCheck by directly assessing visual as well as textualized grounding.

\textbf{Error categories.}
Residual hallucinations mainly involve \emph{fabricated relations}, where no supporting edge exists in $\mathcal{G}_{\mathrm{sub}}$, and \emph{mislocalized entities}, where a plausible predicate is attached to the wrong entity.

\begin{table}[t]
\centering
\caption{Hallucination assessment. Automatic H-Rate is computed with \textsc{MiniCheck}, independently of the revision-time verifier; human evaluation uses 300 balanced examples.}
\label{tab:hallucination_assessment}
\scalebox{0.75}{
\begin{tabular}{|l|c|c|}\hline
\rowcolor{gray!15}
\textbf{Model}
& \textbf{Independent H-Rate (\%)} $\downarrow$
& \textbf{Human Hallucination-Free (\%)} $\uparrow$ \\\hline
LLaVA-1.5~\cite{liu2023llava} & 25.2 $\pm$ 0.8 & 39.5 $\pm$ 1.2 \\
Qwen2.5-VL~\cite{bai2025qwen25vl} & 17.6 $\pm$ 0.6 & 56.2 $\pm$ 1.1 \\
InternVL3~\cite{zhu2025internvl3} & 16.8 $\pm$ 0.5 & 57.9 $\pm$ 1.0 \\
SKURG~\cite{yang2023enhancing} & 20.3 $\pm$ 0.7 & 49.8 $\pm$ 1.1 \\
CRAG~\cite{yan2024crag} & 18.1 $\pm$ 0.6 & 54.7 $\pm$ 1.3 \\
RARR~\cite{gao2023rarr} & 15.2 $\pm$ 0.5 & 59.8 $\pm$ 1.0 \\
Flat-RAG & 14.8 $\pm$ 0.6 & 61.0 $\pm$ 1.1 \\
\textbf{GraphLoom} & \textbf{10.5 $\pm$ 0.4} & \textbf{68.5 $\pm$ 0.9} \\\hline
\end{tabular}
}
\end{table}

Overall, GraphLoom achieves a lower unsupported-claim rate than the evaluated VLM, graph-RAG, corrective-retrieval, and revision baselines. Its improvement over Flat-RAG is consistent with contributions from reliability-calibrated slot routing and evidence-bound revision. Using an automatic evaluator independent of the revision-time verifier, together with human validation, reduces the risk of circular faithfulness evaluation.

\subsection{Implementation Details and Hyperparameter Analysis}
\label{subsec:hyperparam_analysis}

GraphLoom is implemented in PyTorch with Hugging Face Transformers. All backbone models---Qwen3-VL-Instruct, Qwen3-VL-Embedding, and Llama-3.1-8B-Instruct---remain frozen during training. The end-to-end decoder-side trainable modules are the multimodal fusion projection ($\sim$0.8M parameters), \textsc{HieraSlot} projections ($\sim$0.3M), and reliability router ($\sim$0.5M), totaling approximately \textbf{1.6M parameters}. The code is available at\footnote{\url{https://github.com/Zafar-southeast/GraphLoom}}.

\textbf{Training configuration.}
We train with AdamW ($\beta_1=0.9$, $\beta_2=0.95$, weight decay $=0.01$). The learning rate warms up for 1{,}000 steps to $3\times10^{-4}$ and then follows cosine decay to $1\times10^{-6}$. We use batch size 64, gradient accumulation 2, gradient clipping at 1.0, and early stopping on the validation score defined in Section~\ref{sec:training_objective}. Training uses one NVIDIA A100 80GB GPU.

\textbf{Hyperparameters.}
Validation tuning selects $K=15$, two retrieval rounds, $d_{\max}=5$, $N_{\max}=50$, $k=4$, $L_p=32$, $L_{\mathrm{recent}}=64$, $\tau_{\mathrm{edge}}=0.5$, $\tau_{\mathrm{KB}}=0.7$, $\tau_{\mathrm{eval}}=0.6$, and $\tau_{\mathrm{ent}}=0.7$. All values are fixed during testing. Performance is most sensitive to $k$ and $N_{\max}$, which control the trade-off between evidence coverage and computation.

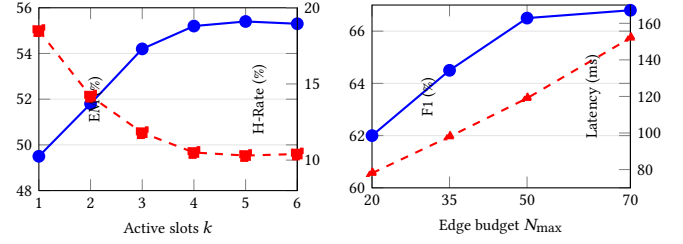
\begin{figure}[!h]
\centering
\footnotesize
\begin{subfigure}{0.48\linewidth}
    \centering
    \begin{tikzpicture}
    \begin{axis}[
            width=5.0cm,
            height=4.0cm,
            xlabel={Active slots $k$},
            ylabel={EM (\%)},
            ylabel style={at={(axis description cs:-0.02,0.5)},anchor=south},
            xmin=1,xmax=6, ymin=48,ymax=56,
            xtick={1,2,3,4,5,6},
            ymajorgrids=true,
            grid style={gray!20},
            ticklabel style={font=\scriptsize},
            label style={font=\scriptsize},
            ylabel style={
                font=\scriptsize,
                yshift=-1.0cm
            }
        ]
        \addplot[blue,mark=*,thick] coordinates {(1,49.5) (2,51.8) (3,54.2) (4,55.2) (5,55.4) (6,55.3)};
    \end{axis}
    
    \begin{axis}[
            width=5.0cm,
            height=4.0cm,
            xmin=1,xmax=6, ymin=8,ymax=20,
            axis y line*=right,
            axis x line=none,
            ylabel={H-Rate (\%)},
            ylabel style={at={(axis description cs:0.8,0.5)},anchor=north},
            ticklabel style={font=\scriptsize},
            label style={font=\scriptsize},
        ]
        \addplot[red,mark=square*,dashed,thick] coordinates {(1,18.5) (2,14.2) (3,11.8) (4,10.5) (5,10.3) (6,10.4)};
    \end{axis}
    \end{tikzpicture}
    \caption{Effect of active slots $k$ on MMQA.}
    \label{fig:slot_k_analysis}
\end{subfigure}
\hfill
    \begin{subfigure}{0.48\linewidth}
    \centering
    \begin{tikzpicture}
    \begin{axis}[
            width=5.0cm,
            height=4.0cm,
            xlabel={Edge budget $N_{\max}$},
            ylabel={F1 (\%)},
            ylabel style={at={(axis description cs:-0.02,0.5)},anchor=south},
            xmin=20,xmax=70, ymin=60,ymax=67,
            xtick={20,35,50,70},
            ymajorgrids=true,
            grid style={gray!20},
            ticklabel style={font=\scriptsize},
            label style={font=\scriptsize},
            ylabel style={
                font=\scriptsize,
                yshift=-1.0cm
            }
        ]
        \addplot[blue,mark=*,thick] coordinates {(20,62.0) (35,64.5) (50,66.5) (70,66.8)};
    \end{axis}
    
    \begin{axis}[
            width=5.0cm,
            height=4.0cm,
            xmin=20,xmax=70, ymin=70,ymax=170,
            axis y line*=right,
            axis x line=none,
            ylabel={Latency (ms)},
            ylabel style={at={(axis description cs:0.8,0.5)},anchor=north},
            ticklabel style={font=\scriptsize},
            label style={font=\scriptsize},
        ]
        \addplot[red,mark=triangle*,dashed,thick] coordinates {(20,78) (35,98) (50,119) (70,152)};
    \end{axis}
    \end{tikzpicture}
    \caption{Effect of edge budget $N_{\max}$ on MMQA.}
    \label{fig:budget_analysis}
\end{subfigure}

\caption{Sensitivity analysis of active slots and edge budget. The selected defaults ($k=4$, $N_{\max}=50$) balance answer quality and computation.}
\label{fig:hyperparam_analysis}
\end{figure}

\subsection{Latency and Efficiency Breakdown}
\label{subsec:latency_efficiency}

We report full-pipeline and online latency to separate cacheable graph construction from query-time evidence use. Full-pipeline latency includes scene description generation, triple extraction, entity linking, KG enrichment, initial and optional interleaved/corrective retrieval, decoding, verification, and revision. Online latency assumes cached instance-level graph construction and includes query-conditioned retrieval, slot routing, decoding, and conditional correction/revision. Conditional modules are reported as amortized per-query costs over the evaluation set, with non-triggered cases contributing zero additional cost.

At inference, GraphLoom retrieves 38.2 triples per query on average, activates at most $|\mathcal{A}_t|\leq4$ graph slots per decoding step, and retains $|I_t^{(\ell)}|\leq88$ sparse self-KV tokens. At $t=512$, this reduces per-step attention FLOPs by approximately 76\% relative to full causal attention. Table~\ref{tab:latency_breakdown} reports the module-level latency breakdown.

\begin{table}[!h]
\centering
\caption{Latency breakdown of GraphLoom on one NVIDIA A100 80GB GPU. Conditional modules are reported as amortized per-query costs.}
\label{tab:latency_breakdown}
\scalebox{0.88}{
\begin{tabular}{|l|c|}\hline
      \rowcolor{gray!15}
      \textbf{Module} 
    & \textbf{Latency / query (ms)} $\downarrow$ \\\hline
        Scene description generation & 142 \\
        Triple extraction & 44 \\
        Entity linking + KG enrichment & 18 \\
        Initial retrieval & 9 \\
        Interleaved retrieval & 13 \\
        Corrective retrieval & 7 \\
        Slot routing + \textsc{KG-JSA++} decoding & 66 \\
        Verification--revision & 24 \\\hline
        Full-pipeline latency & 323 \\
        Online latency with cached graph & 119 \\\hline
\end{tabular}
}
\end{table}

Most full-pipeline cost comes from cacheable graph construction, particularly scene description generation and triple extraction. With cached instance-level graphs, online latency is 119\,ms per query, dominated by slot routing, \textsc{KG-JSA++} decoding, and verification--revision. This separation clarifies the cost of online evidence use while showing that decoder-time attention remains controlled by the active-slot and sparse self-KV budgets.

\section{Conclusion and Future Work}
\label{sec:conclusion}

We presented \textbf{GraphLoom}, a reliability-calibrated multimodal KG-RAG framework that retrieves compact evidence subgraphs and injects selected graph memories into a frozen decoder through \textsc{HieraSlot} and \textsc{KG-JSA++}. Experiments on ScienceQA, MultiModalQA, and OK-VQA show consistent gains in answer quality and faithfulness, with additional improvements in MMQA retrieval. Scalability, robustness, and latency analyses further demonstrate stable performance under noisy evidence pools and moderate threshold variation. These results suggest that reliability-aware evidence routing can provide an effective alternative to larger models or longer contexts. Future work will explore adaptive retrieval, stronger multimodal indexing, and extensions to document, chart, table, and video QA.

\section*{GenAI Usage Disclosure}

Generative AI components used within GraphLoom are described in the methodology and experimental setup. OpenAI ChatGPT was additionally used to assist with code generation, debugging, experimental output preparation, LaTeX, language refinement, and consistency checking. All generated code and reported outputs were reviewed and validated by the authors, who made the methodological and technical decisions and take full responsibility for the final work.

\begin{acks}
This work was supported by the NSFC under Grant No. 6509009704.
\end{acks}

\bibliographystyle{ACM-Reference-Format}
\bibliography{graphloom_references}

\end{document}